\pdfoutput=1

\documentclass[11pt]{article}

\usepackage[]{acl}
\usepackage{graphicx}
\usepackage{times}
\usepackage{latexsym}

\usepackage[T1]{fontenc}

\usepackage[utf8]{inputenc}

\usepackage{microtype}

    \title{Detecting Rumor Veracity with Only Textual Information by Double-Channel Structure}

\author{Alex Gunwoo Kim$^*$ \\
  Seoul National University \\
  \texttt{kimgunwoo95@snu.ac.kr} \\\And
  Sangwon Yoon$^*$ \\
  Artificial Society Inc.\\
  \texttt{sangwon38383@snu.ac.kr} \\}

\begin{document}
\maketitle
\def\thefootnote{*}\footnotetext{Equal contribution.}\def\thefootnote{\arabic{footnote}}
\begin{abstract}
\citet{kyle1985} proposes two types of rumors: informed rumors that are based on some private information and uninformed rumors that  are not based on any information (i.e. bluffing). Also, prior studies find that when people have credible source of information, they are likely to use a more confident textual tone in their spreading of rumors. Motivated by these theoretical findings, we propose a double-channel structure to determine the ex-ante veracity of rumors on social media. Our ultimate goal is to classify each rumor into true, false, or unverifiable category. We first assign each text into either certain (informed rumor) or uncertain (uninformed rumor) category. Then, we apply lie detection algorithm to informed rumors and thread-reply agreement detection algorithm to uninformed rumors. Using the dataset of SemEval 2019 Task 7, which requires ex-ante threefold classification (true, false, or unverifiable) of social media rumors, our model yields a macro-F1 score of 0.4027, outperforming all the baseline models and the second-place winner (\citealp{gorrell2019}). Furthermore, we empirically validate that the double-channel structure outperforms single-channel structures which use either lie detection or agreement detection algorithm to all posts.\footnote{The code to replicate the results of this article can be found here: \url{https://github.com/swarso95/rumour_analysis-}.}
\end{abstract}

\section{Introduction}

Detecting the veracity of rumors spreading out on various social media platforms has been of great importance. Indeed, several studies find that online rumors can affect human behaviors (\citealp{pound1990}; \citealp{jia2020}). However, detecting the veracity of rumors is not a simple task. Unlike news articles which are considered \textit{ex-post}, rumors are \textit{ex-ante} (\citealp{vosoughi2018}; \citealp{shu2017a}). At the time when a rumor originates, the information user is not able to determine its veracity by checking whether the event has happened or not. Instead, the user can make his best guess based on the information set that he has been exposed to. In contrast, we can check the veracity of a news article immediately by comparing it with the event that the article is referring to (\citealp{cao2018}). There can be diverse definitions of rumors, but in our study we define the rumors as "\textit{information that cannot be verified at the time of origination} (\citealp{gorrell2019})".\footnote{This definition excludes tasks such as PHEME from our scope of analysis since they require "fact-checking" instead of "ex-ante prediction of veracity."} Therefore, whether a rumor is false or not can only be determined afterward when the user can objectively observe the event (\citealp{zubiaga2016}). 

In our research, we use only the textual features of the posts and their corresponding replies, mitigating the concern that our results are driven by external information that was not readily available to the general public at the early stage of rumor origination. Also, our model shows that textual features embedded in social media posts can reasonably predict the ex-ante veracity of rumors.

\begin{figure}[t]
\centering
\includegraphics[width = 1.0\linewidth ]{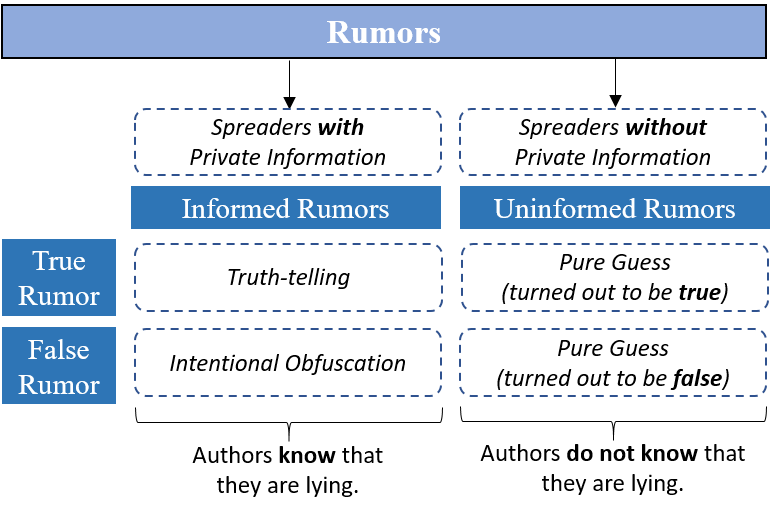}
\caption{This figure illustrates the conceptual classification of rumors based on prior linguistics literature. Our model motivates from these two different subgroups.}
\label{class}
\end{figure}

\citet{kyle1985} provides a theoretical model that explains the motivation of spreading rumors. The model includes two types of rumor spreading: (i) rumors based on private information and (ii) rumors not based on any information (i.e. bluffing). Spreaders with private information can either deliver the correct information that they have or intentionally distort the information. On the other hand, there can be spreaders without private information. They take advantage of their social influence and spread some made-up rumors in favor of their benefits (\citealp{van2003}). Refer to Figure~\ref{class} for the visual representation of rumor classification.

Studies on linguistics find that the perceived credibility of information source affects the tone of rumors on social media (\citealp{kim2019}; \citealp{kamins1997}; \citealp{difonzo2010}). The more credible the information source is, the more confident the textual tone is. For instance, rumors based on concrete source of information are likely to include a reference link or refer to specific identities. In contrast, bluffing is less likely to encompass the source of information.

Combining these two lines of literature, rumors based on private information and rumors \textit{not} based on private information are systematically and linguistically different. However, prior studies that intent to identify the “ex-ante” veracity of social media rumors (e.g. \citealp{enayet2017}; \citealp{wu2015}; \citealp{rao2021}) treat every rumor equally. In other words, they apply the same logic or algorithm to both types of rumors. To tackle this issue, we conjecture that dividing the sample into “informed rumors” (rumors that are based on private information) and “uninformed rumors” (rumors that do not have any information background) and applying different algorithms to the two subgroups can improve the performance of veracity detection. 

Motivated by the linguistic differences between the two rumor types, we first divide the sample based on the textual confidence of rumor texts. This algorithm classifies each rumor into certain (informed rumors) or uncertain (uninformed rumors) category. As in \citet{kyle1985}, informed spreaders can strategically choose whether or not to truthfully report the private information that they have. If they choose to distort the information, the spreaders are intentionally lying. In contrast, they might opt for truth-telling. Therefore, we apply the lie detection algorithm to informed rumors to determine their ex-ante veracity.

On the other hand, for uninformed rumors, the spreaders are not intentionally lying nor are they truthfully reporting. Therefore, we do not expect lie detection algorithm to function properly. Instead, we rely on the agreement detector algorithm (\citealp{kumar2019}; \citealp{yu2020}). Prior literature finds that when primary replies are generally in accordance with the original thread, the thread is likely to be true ex-post, and vice-versa (\citealp{akhtar2018}). In our model, we use primary replies and calculate their agreement scores with the main thread. The logic beyond this algorithm is that the wisdom of the crowd plays a role in social media platforms to provide accurate information (\citealp{brown2019}; \citealp{yu2020}). We leave the mathematical details for Sections 3.1 and 3.2.

In our study, we further validate this theory-motivated double-channel approach by showing that our model outperforms the single channel structures (applying lie detection algorithm or agreement detection algorithm to both channels). Section 4.1 outlines the relative performance of double-channel model compared with other structures and with other competing models of SemEval 2019. Specifically, our model achieves a macro-F1 score of 0.4027, which is approximately 12\% points higher than that of the second-place winner. 

\begin{figure}[t]
\centering
\includegraphics[width = 1.0\linewidth ]{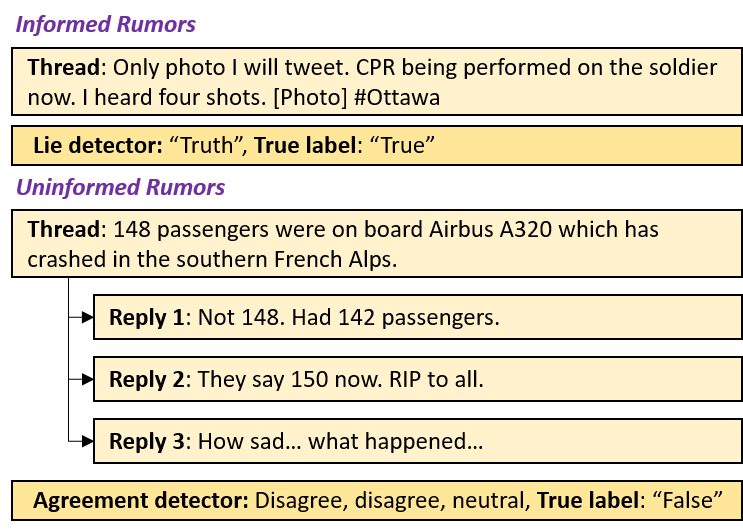}
\caption{This figure illustrates an example of the classification results of our model.}
\label{example}
\end{figure}

Figure~\ref{example} provides an example of the classification results of our model. The uninformed thread does not refer to any source information while the informed one does so. Lie detection algorithm correctly classifies the veracity of the informed rumor. On the other hand, agreement detector captures whether each primary reply is in accordance with the main thread. The algorithm correctly classifies the thread to be false.

Our research contributes to the existing line of literature for at least two reasons. First, we are the first to employ a double-channel model to detect the veracity of rumors. This approach reflects the rumor classification (informed and uninformed) proposed by the linguistics literature. We show that the lie detection algorithm is relatively more appropriate for classifying informed rumors and that the agreement detection is more accurate when classifying uninformed rumors. After employing a BERT-based certainty classifier to divide the samples into two subgroups, we find a significant increase in our classification accuracy.

Second, we also use minimal information to obtain our results. Our F1 score falls behind the winner of SemEval 2019 Task 7, primarily due to the scope of the information that we use. The winner exploits a variety of peripheral information such as the account credibility or the number of followers (\citealp{li2019}), which explains a great portion of their results. However, such a model cannot be applied to anonymous rumors or rumors posted by relatively "new" users. In contrast, our model operates even without considering the peripheral or user-specific information, allowing it be applied to even anonymous rumors in social media. Also, since the second-place winner primarily focuses on the textual dimension of Twitter posts, we find the second-place winner more comparable to our assumptions and experiments. 

\section{Related Works}
\subsection{Information Sets}
Prior literature mainly relies on two information sets to calculate the ex-ante veracity of rumors. First, several studies use user information such as the number of followers, the number of replies, the existence of hashtags and photos, and the number of previous tweets to determine the veracity of each rumor (\citealp{castillo2011}; \citealp{vosoughi2015}; \citealp{liu2018}; \citealp{li2019}). This line of research assumes that the users who care about their accounts' reputation are likely to post true rumors. However, it is difficult to measure the account's credibility when the rumor originates since the account information is time-variant. Even though a specific account currently has many followers, we cannot guarantee that the account used to have the same number of followers when the rumor originated. Furthermore, such information is not available for anonymous rumors.

Second, several studies apply linguistic features to detect false rumors. Some studies measure the subjectivity of the posts using some attribute-based textual elements such as subjective verbs and imperative tenses (\citealp{li2019}; \citealp{ma2017}; \citealp{liu2015}). \citet{vosoughi2015} analyzes the sentiment of tweets under various circumstances and classify the tweets using the contextual information. \citet{barsever2020} develop a better-performing lie detector with BERT, indicating that unsupervised learning can outperform traditional rule-based lie detection algorithms. However, the linguistic feature-based approach has limitations in that most of the rumors are arbitrary in nature, and lie detection, which is based on the author's intention, may not function well in the domain that contains many random posts.

Other research focuses on the network model to capture information propagation (\citealp{gupta2012}; \citealp{rosenfeld2020}). Also, \citet{liu2018} develop a model that examines the early detection of rumors with RNN classification. Also, several works aim to determine whether a given online post is a rumor or not (\citealp{kochkina2018}) by implementing a multi-task learning algorithm.

\subsection{Classification Algorithm}
While several studies deal with improving the input dataset, others focus on improving the classification algorithm. Some early studies are based on Support Vector Machine (SVM) (\citealp{enayet2017}; \citealp{wu2015}) or neural networks (NN) to conduct the classification (\citealp{ma2017}; \citealp{wang2018}). 

Recent works turn to unsupervised learning of rumors. Instead of inputting a number of user-specific variables, \citet{rao2021} develop STANKER, a fine-tuned BERT model which incorporates both the textual features of posts and their comments. This model inputs comments as one of the crucial auxiliary factors, measuring the co-attention between the posts and comments. Our model differs from STANKER for at least two reasons. First, unlike STANKER which uses single-channel approach, we design a double-channel approach. This approach allows us to apply a more appropriate classifier to each thread. Second, STANKER is trained with more than 5,000 labeled observations. These observations do not include the "unverified" category as well. However, since our train set contains only 365 observations with three different labels, we utilize external open-source datasets from similar (yet slightly different) domains to further train each phase of our model. Therefore, we aim to improve the performance of the model with the minimal information and fine-tune the model to mitigate the domain-shift problem.

On the other hand, \citet{yu2020} develops a Hierarchical Transformer which disaggregates a thread into subthreads. Then, they process the stance labels obtained from the subthreads to determine the veracity of a rumor. Their method focuses on the mutual interaction among the users but may not function properly at the early stage of rumor origination when there are not enough reply posts. Furthermore, \citet{dougrez2021} employ a Variational Autoencoder to filter out the topics that are useful in stance determination and achieve a macro-F1 score of 0.434 on PHEME dataset. 

\section{Model Design}
\subsection{Overall Structure}
Our model is the first to introduce a double-channel approach in rumor veracity detection. We first divide the sample into two subsamples depending on the textual confidence of each thread. Here, a confidence score examines whether the author is writing the post with a strong belief or not (\citealp{farkas2010}). Authors who spread informed rumors are more likely to be confident in their postings (\citealp{difonzo2010}). Therefore, our BERT-based uncertainty-classifier assigns each thread into one of the two categories: certain (informed rumor) and uncertain (uninformed rumor) (\citealp{devlin2018}). We assume that informed rumors are based on educated belief, insider information, or other reliable sources. We name this step Phase 1.

Then, we turn to lie detection algorithms for informed rumors. Note that when the author has baseline information, it is the author's choice to decide whether or not to disclose the true information to the public. Textual lie detection focuses on lexical cues that are prevalent in intentional lies (\citealp{masip2012}) and examines the author's intention – it identifies whether the writer is intentionally distorting actual information. If the authors decide to distort the information, the lie detector is expected to identify such intention (\citealp{mansbach2021}; \citealp{barsever2020}). We use a BERT-based lie classifier to assign the threads into a true or false category. We call this step Phase 2-1.

On the other hand, for uninformed rumors, we cannot rely on the linguistic lie detection. Uninformed rumors are written by people who do not have any specific reference when spreading the rumors. In other words, they make an uninformed guess or even write some random facts in their accounts. Since the writers do not intend to deceive other people (they do not even know what is true or false), the lie detection algorithm may not function properly. Therefore, we should take a different approach to determine the veracity of such rumors. Here, we focus on the agreement score of each reply. Users actively respond to the rumors in social media, and the wisdom of the crowd is known to generate remarkably accurate information (\citealp{brown2019}; \citealp{navajas2018}). In our study, we calculate the degree of agreement of each primary reply to the thread. Then, using the agreement score of the replies, we estimate the veracity of the thread. We call this step Phase 2-2.

\begin{figure}[t]
\centering
\includegraphics[width = 1.0\linewidth ]{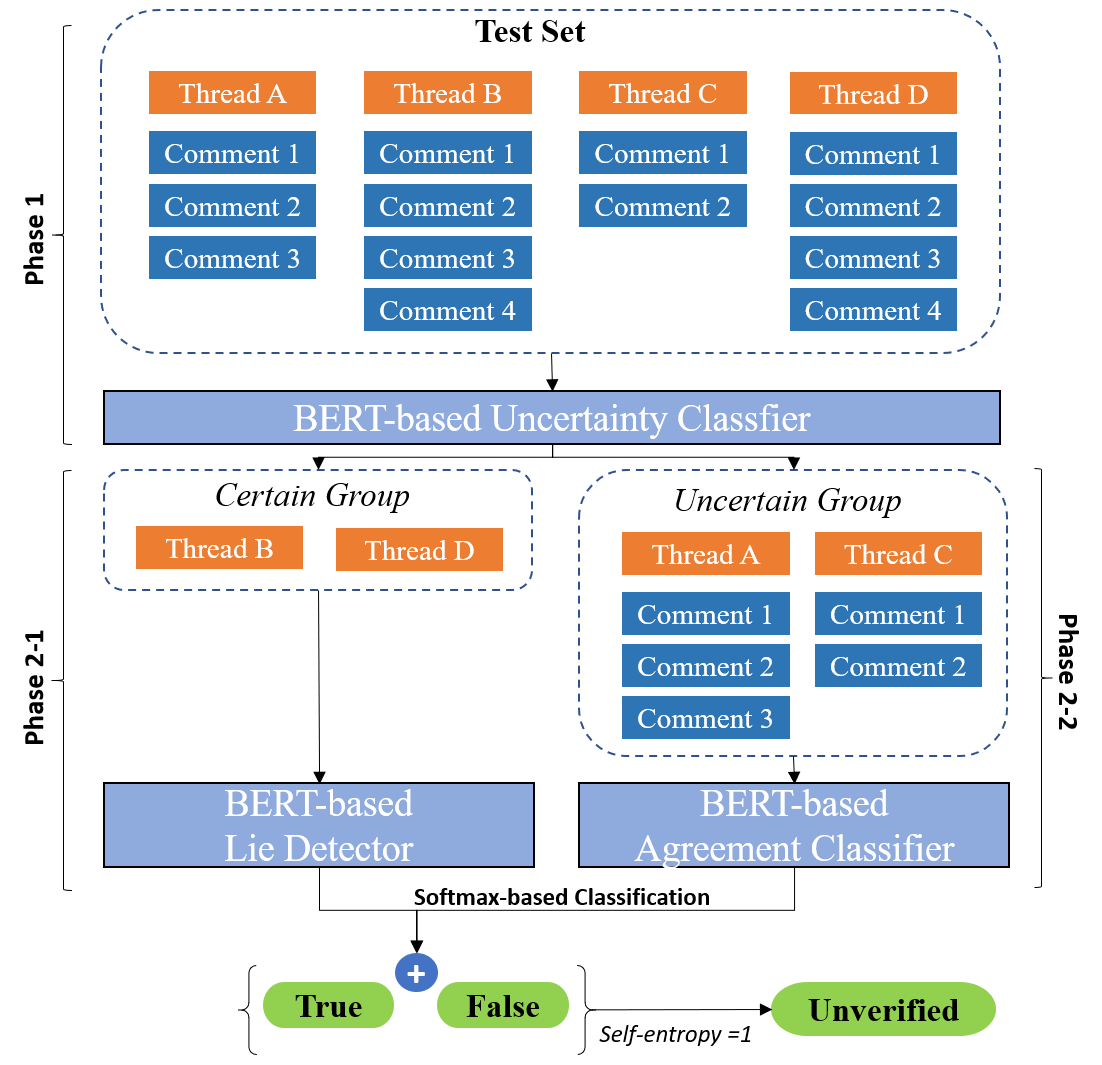}
\caption{This figure illustrates the model pipeline. Uncertainty classifier (Phase 1) divides the sample into two subgroups, and lie detector (Phase 2-1) and agreement classifier (Phase 2-2) further classifies each thread into true or false category. We assign the observations with self-entropy of 1 to unverified category.}
\label{pipe}
\end{figure}

For the visual representation of our pipeline, refer to Figure~\ref{pipe}. We use Tesla V100 SXM2 32GB GPU to train our model. We use BERT in all phases of our model since BERT and its variants achieve the state-of-the-art performance in text classification tasks (\citealp{liu2019}; \citealp{lan2019}).
\subsection{Phase 1: Detecting Linguistic Certainty}
We develop a BERT-based certainty classifier. Our classifier is a binary classifier based on a BERT sentence classifier. Our goal is to assign each sentence (Twitter or Reddit thread) into one of the two categories: certain or uncertain. We first train our model with the labeled dataset provided in CoNLL-2010 Shared Task (\citealp{farkas2010}). The dataset contains binary labels (certain or uncertain) and 7,363 observations. We use a batch size of 32 and a learning rate of 5e-5. We train the model for five epochs and use Adam optimizer.

We apply the trained BERT classifier to our train set. This process yields 365 distinct thread-label pairs. However, the domain of the dataset that we use to train the model slightly differs from the domain of the dataset that we have. To tackle this domain-shift issue, we sample 21 observations from each category (certain and uncertain) and re-train the model for five epochs. We select the same number of observations from the two categories to mitigate the concern arising from severely imbalanced classifications. We use a batch size of 32 and a learning rate of 5e-5. This procedure assuages the potential bias due to domain-shifting.
	
We set a label smoothing rate of 0.2 for both training steps. Label smoothing resolves the classification imbalance due to the differences in the two domains and the potential overfitting due to the limited number of our training samples (\citealp{szegedy2016}). We apply Phase 1 to all test samples and obtain 81 distinct thread-label pairs. 17 of them are classified as informed rumors, and the remaining 64 observations are classified as uninformed rumors.

\subsection{Phase 2-1: Fake Rumor Identification with Lie Detection Algorithm}
We apply Phase 2-1 to informed rumors from Phase 1. We develop a BERT-based binary sentence classifier to detect lies from lexical cues. Similarly, we take a two-step approach to train the model. First, we use the open-source dataset to train a model that detects scams and lies in social media (\citealp{ott2011}; \citealp{ott2013}). This dataset contains 1,600 pre-labeled texts. We train the model for five epochs with a batch size of 32, a learning rate of 5e-5, and a label smoothing rate of 0.3. We also use Adam optimizer.

Then, we fine-tune the model with the train dataset of SemEval 2019 Task 7. According to the definition, unverified samples are those with zero confidence scores. Therefore, when fine-tuning our model, unverified observations are of no use. We exclude the unverified samples and use only observations with true or false labels. We fine-tune the model for one epoch using the samples that are classified as certain in Phase 1. Our batch size is 32 and learning rate is 5e-5. Unlike certainty classification of Phase 1, the domains and objectives of the external dataset that we use are similar to our primary goal – determining the veracity of a given statement. However, in Phase 1, the surrogate dataset aims at discerning non-factual and factual information. That is, the objectives of the two tasks are similar but not the same. Therefore, we train the model for five epochs in Phase 1. In Phase 2-1, since the two tasks deal with the same agenda, it suffices to fine-tune the model for one epoch.	

When applied to the test set, our lie detector yields 81 distinct thread-label pairs. The label includes true and false indicators based on the softmax values. That is, when the softmax value of true is larger than the softmax of false the program returns true and vice versa. Following the definition of the unverified rumors, we classify the samples with self-entropy score of 1 into unverified category. Otherwise, we use the labels obtained from our lie detector.

The self-entropy of each observation is 
$$H(x) = -\frac{1}{\log 2}\sum_{n=0}^1l_n(x)\log l_n(x)$$
, where $x$ denotes each observation and $l_n(x)$ denotes the probability that $x$ belongs to each category ($n = 0, 1$).
\begin{figure*}[t]
\centering
\includegraphics[width = 1.0\linewidth ]{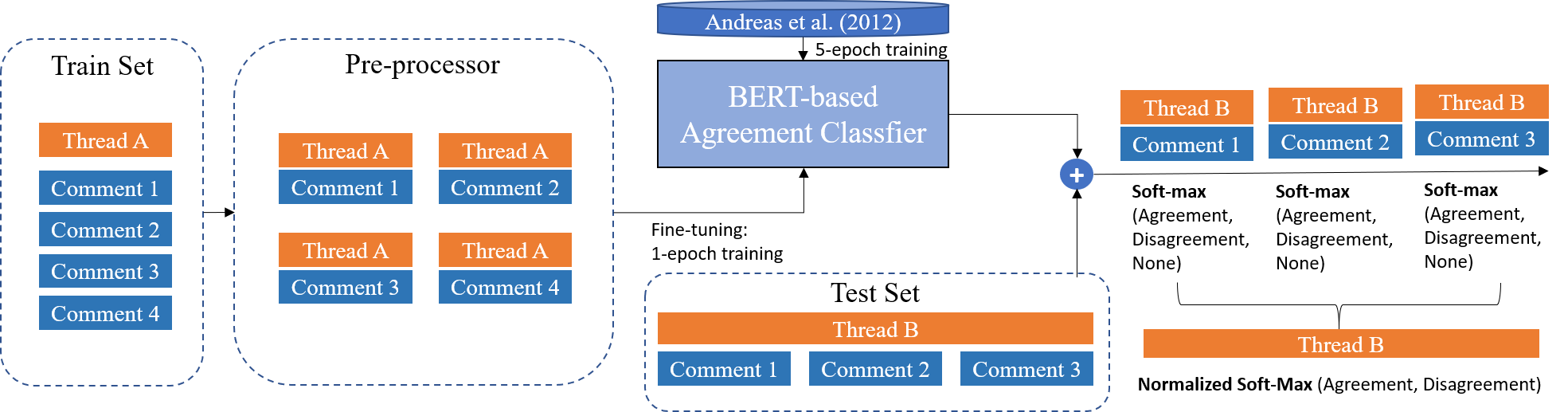}
\caption{This figure illustrates the pipeline of Phase 2-2. We pre-train the BERT model with the dataset provided by \citet{andreas2012} and fine-tune the model with pre-processed train set of SemEval 2019 Task 7. Then we apply the BERT-based agreement detector to thread-reply pair of the test set and obtain soft-max value vectors. We discard the soft-max values of \textit{none} since \textit{none} does not provide additional information about the veracity of the rumors.}
\label{phase2-2}
\end{figure*}
\subsection{Phase 2-2: Fake Rumor Identification with Reply Agreement Score}
We apply Phase 2-2 to uninformed rumors from Phase 1. Here, we develop a BERT-based triple sentence classifier that assigns each sentence pair into one of the three categories: agreement, disagreement, and none. Here, the input is a sentence pair composed of one thread and its corresponding primary reply. For instance, in Figure~\ref{phase2-2}, since thread A has four primary replies, we construct four sentence pairs. We exclude non-primary replies (replies to the previous replies) since it is unclear whether such non-primary replies are agreeing (or disagreeing) to the thread itself or to the primary reply. Therefore, the classifier measures whether the primary reply is in accordance with the thread or not. We also take a two-step approach to train the model.

First, we train the BERT-based triple classifier with an open-source dataset (\citealp{andreas2012}). The dataset contains 1,163 sentence pairs with agreement labels. Specifically, it includes 609 agreement pairs and 554 disagreement pairs. We train the model for five epochs with a batch size of 32, a learning rate of 5e-5, and a label smoothing rate of 0.3. We also use Adam optimizer.

Then, we fine-tune the model with the train set of SemEval 2019 Task 7. We filter out primary responses from the dataset and create thread-reply pairs. We label the pairs with the labels pre-assigned to each thread. This process yields 2,372 distinct thread-reply pairs. Then we train the model for one epoch with batch size 32 and learning rate 5e-5. The task of \citet{andreas2012} aims at determining whether each reply is in accordance with the thread, which is identical to our objective. Hence, we fine-tune the model for one epoch.

Applying the classifier to uninformed rumors yields the softmax values for (agreement, disagreement, none). We discard the softmax value of none and sum the softmax values of agreement and disagreement for each thread. Then, we normalize the values so that they sum up to be one. As in Phase 2-1, the program returns true when the softmax value of the agreement is larger than that of disagreement and vice versa. 

For a formal representation, let $X_i$ denote the thread and $y_m^i$ denote the $m$th primary reply to $X_i$. Suppose that we have $k$ threads and $n_i$ ($i$ is an integer between 1 and $k$) is the number of primary comments corresponding to $X_i$. We form up the pairs $(X_1, y_1^1), \cdots, (X_1, y_{n_1}^1), \cdots, (X_k, y_1^k), \cdots, \\(X_k, y_{n_k}^k)$. BERT model returns a softmax vector of each pair $(a_l, b_l, c_l)$, where $(a,b,c)$ denotes the softmax vector of (agreement, disagreement, none). We obtain $\sum_{i=1}^kn_i$ softmax vectors. Then, for $X_i$, we sum up the softmax values to obtain the normalized softmax vector.
$$
\left(\frac{\sum_{k=1}^{n_i}a_k}{\sum_{k=1}^{n_i}a_k+\sum_{k=1}^{n_i}b_k}, \frac{\sum_{k=1}^{n_i}b_k}{\sum_{k=1}^{n_i}a_k+\sum_{k=1}^{n_i}b_k} \right)
$$

If the first softmax is larger than the second, we classify $X_i$ to be true. If the second softmax is larger than the first, we classify $X_i$ to be false.

\begin{table*}
\centering
\begin{tabular}{|l|l|l|l|l|}
\hline
                                    & Macro-F1 & Accuracy & Precision & Recall\\ \hline
\textbf{Double-Channel}                      & \textbf{0.4027}         & \textbf{0.4938 }  & \textbf{0.5064}& \textbf{0.4043}\\ \hline
Single-Channel (Lie Detector)       & 0.3447         & 0.4444  & 0.3362 & 0.3706\\ \hline
Single-Channel (Agreement Detector) & 0.3668         & 0.4444  & 0.4813 & 0.3700\\ \hline
Double-Channel with Inverse Detectors & 0.3145         & 0.3567  & 0.2981 & 0.3374 \\ \hline
Baseline (LSTM)                     & 0.3364         & -    & -&-    \\ \hline
Baseline (NileTMRG)                 & 0.3089         & -    & -&-    \\ \hline
Baseline (Majority class)           & 0.2241         & -    & -&-    \\ \hline
WeST (CLEARumor)                    & 0.2856         & -    & -&-    \\ \hline
eventAI                             & 0.5770         & -    & 0.5960& 0.6030    \\ \hline
\end{tabular}
\caption{\label{results}
This table demonstrates the relative performances of the models that we develop, the baseline models of SemEval 2019 Task 7, and the second-place winner of the task (WeST). Single-channel models include the model that applies lie detector to all observations and the model that applies agreement detector to all observations. Double-channel model with inverse detectors apply lie detection algorithm to uncertain group (uninformed rumors) and agreement detection algorithm to certain group (informed rumors).
}
\end{table*}

Also, we assign the observations with the self-entropy value of 1 to the unverified category. We calculate the self-entropy using the same formula with Phase 2-1.

We discard the softmax values of none because replies that do not fall under either agreement or disagreement category do not have informational value. By allowing the none category and discarding the none category samples, we aim to deliberately examine the replies' intent (\citealp{li2019}). Refer to Figure~\ref{phase2-2} for the graphical illustration of Phase 2-2.

\subsection{Data and Pre-processing}
Our primary input data is the open-source data released in SemEval 2019 Task 7. Specifically, we aim to improve the model performance of Task 7B, in which the participants are asked to classify each rumor into one of the three categories (true, false or unverifiable). The dataset contains 365 train set observations. Each observation consists of one thread (Twitter or Reddit) post and its corresponding replies. Replies include the primary replies (replies that respond directly to the main post) and secondary replies (replies that respond to other replies). In our task, we do not use replies other than primary replies. We first retrieve all main posts (threads) from the dataset. The threads often include hashtags or web addresses starting with http. Several studies including \citet{li2019} use this as auxiliary information in their analysis - they include an indicator variable that equals one when the thread has a hashtag or web address inside. However, in our research, we focus only on textual features and do not need such information. Further, given that the threads are relatively short, uninterpretable hashtags or web addresses might distort the results. Hence, we delete all hashtags and web addresses that start with "http".

Then, we turn to the comments. The dataset contains a structure file in json format for each thread. The structure file explains the format of each thread such as how many comments are there, the time when each comment is posted, the ID of the author and the ID of the comment. From the json file, we identify the primary comments and pair them with their corresponding thread. We also cleanse the texts by removing all the hashtags and web addresses.

\section{Results}
We present our results in Table~\ref{results}. We report two main evaluation metrics, macro-F1 and accuracy, and two supplementary metrics, precision and recall. Macro-F1 is the harmonic average of the precision and recall ratios, while accuracy is the ratio of correct classifications to the total number of observations. 

\subsection{Justification of Double-Channel Structure}
In support of our conjecture, we re-train the Phase 2-1 and Phase 2-2 classifiers with all observations, and report the results when the classifiers are applied to all posts without the certainty classification. The results yield the macro-F1 scores of 0.3447 and 0.3668, respectively. Additionally, we also report the prediction accuracy when lie detection algorithm is applied to uninformed rumors and agreement detection algorithm is applied to informed rumors. The macro-F1 score and accuracy (0.3145 and 0.3567) become even lower. As clearly indicated, dividing the total sample into two subgroups significantly improves the classification performance. This improvement is primarily because each classifier is applied to the observations that the classifier is intended to function well. These empirical results further validate our novel double-channel structure along with its theoretical background.

\subsection{Overall Performance}
Our double-channel model achieves a macro-F1 score of 0.4027 and an accuracy of 0.4938. In terms of precision and recall, it achieves 0.5064 and 0.4043, respectively. \footnote{The model correctly classifies 19 true rumors out of 31, 20 false rumors out of 40, and 1 unverified rumor out of 10.} This model outperforms all the baseline models proposed in SemEval 2019 Task 7 and the model developed by the second-place winner. Note that our program only refers to textual information of the main threads and their primary replies. We intentionally do not include user-specific peripheral information to demonstrate that the double-channel approach can significantly improve the classification outcomes.

Our model outperforms the second-best program (WeST) by approximately 12\% points in terms of macro-F1. With the double-channel classification system that we develop, we manage to accurately classify false rumors at their early stage, without considering the peripheral information sets. Our model falls behind the winner of SemEval 2019 Task 7, primarily because we use limited scope of information. We intentionally discard all other information but textual information of the threads and their primary replies. In contrast, the winner exploits a wide variety of information such as account credibility and the existence of hashtags. Unlike the winner, our program can be applied to anonymous rumors without any clue about the author information.

\subsection{Some Restrictions on Replies (Phase 2-2)}
In our main model, we use all primary replies to the main threads, regardless of their dates created. However, we acknowledge that if it takes too much time to collect the reply data, our model cannot calculate the veracity in a timely manner. Since early veracity detection is one of our main contributions, we restrict the replies to be posted within 1-, 3-, and 5-day period from the original thread. Table~\ref{reply} reports the results.

As we restrict the replies to be posted within 1 day from the original thread, we lose 3 threads. Furthermore, we experience a slight decrease in our predictive accuracy and macro-F1 score. However, as we loosen our restriction from 1-day window to 5-day window, we observe a gradual restoration in both accuracy and macro-F1. In summary, our model reasonably predicts the veracity of rumors even in a 1-day window from the origination of rumors and it gradually becomes more accurate in a 5-day window. Note that the average number of replies is 11.96 even when we restrict our window to 1-day period, allowing us to have enough replies to expect the effect of the wisdom of the crowd.\footnote{To further validate this argument, we repeat the same exercise after excluding the threads with only one reply in 1-day restriction sample and achieve a macro-F1 of 0.3570 and accuracy of 0.4800. When we exclude threads with less than 3 replies, we achieve a macro-F1 of 0.3637 and accuracy of 0.4857.}

\begin{table}
\centering
\begin{tabular}{|c|l|l|l|l|}
\hline
         & \multicolumn{1}{c|}{F1} & \multicolumn{1}{c|}{Accuracy} & \multicolumn{1}{c|}{Avg \#} & \multicolumn{1}{c|}{\# thr} \\ \hline
Original & 0.4027                  & 0.4938                        & 14.96                                                                                                                                                                                                               & 81                           \\ \hline
1-Day    & 0.3418                        &  0.4743                             &      11.96                                                                                                                                                                                                                                                                                                                & 78                            \\ \hline
3-Day    & 0.3542                        &  0.4815                             &    14.37                                                                                                                                                                                                                                                                                 &  81                           \\ \hline
5-Day    & 0.3827                        &  0.4938                             &    14.58                         &  81                           \\ \hline
\end{tabular}
\caption{\label{reply}
Avg \# denotes the average number of replies and \# thr denotes the number of distinct threads. $n$-Day denotes the sample when we restrict the replies to be posted within $n$ days from the original thread ($n$=1,3,5).}
\end{table}

\section{Conclusion}
Perfectly determining the veracity of rumors at the time of their origination is impossible. Nonetheless, an increasing number of rumors are spreading out via social media, and people are affected by those rumors. Therefore, sorting out the "likely-fraudulent" rumors at their early stage is of great importance to information users.

Our model takes minimal textual information and achieves a reasonable prediction accuracy in the SemEval 2019 Task 7 dataset. This dataset contains only 365 train samples and 81 test samples, but requires three-way classification. We achieve the macro-F1 score of 0.4027 in this task, which is approximately 12\% points higher than that of the second-place winner which also focuses on the textual features of posts. 

Instead of integrating a wide variety of user-specific information, our model shows that textual features have sufficient predictive power in determining the veracity of rumors. More importantly, we demonstrate that applying a uniform classifier to all Twitter and Reddit posts can harm the model's performance. Instead, we apply a double-channel approach in rumor veracity detection. We divide the sample into two subgroups depending on the textual certainty and apply two different classifiers to each subgroup. Also, by using only textual features of a post and its primary replies, this study responds to \citet{li2019survey}'s call for research that enables the early detection of rumor veracity.

Our research can be successfully implemented in the real world setting. Our model, which does not rely on user-specific information (e.g. the number of followers, the number of previous posts, etc.), can even be implemented to determine the veracity of \textbf{anonymous} rumors. The model produces a rapid veracity prediction. That is, we can produce the results almost immediately for informed rumors and within several days for uninformed rumors. Ultimately, providing users with predicted veracity information can help their potential decision making.

\section{Acknowledgement}
The authors deeply appreciate helpful comments from Bok Baik. Furthermore, the authors appreciate the GPU support from Artificial Society. 

\bibliography{anthology,custom}
\bibliographystyle{acl_natbib}

\end{document}